\documentclass[11pt]{article}

\usepackage[final]{acl}
\usepackage{times}
\usepackage{latexsym}

\usepackage[T1]{fontenc}
\usepackage[utf8]{inputenc}

\usepackage{microtype}

\usepackage{inconsolata}

\usepackage{graphicx}

\usepackage{amsmath}
\usepackage{amssymb}
\usepackage{mathtools}
\usepackage{bm}              % for \bm{} bold math symbols like \bm{\ell}

\usepackage{booktabs}        % for \toprule, \midrule, \bottomrule
\usepackage{multirow}        % for \multirow in tables

\usepackage{graphicx}        % for \resizebox
\usepackage{xcolor}
\usepackage{tcolorbox}       % for the prompt boxes in §2
\tcbuselibrary{skins, breakable}

\usepackage{tikz}
\usetikzlibrary{shapes.geometric, arrows.meta, positioning, calc, shadows}

\usepackage[capitalize,noabbrev]{cleveref}

\definecolor{darkred}{RGB}{150,0,0}
\newtcolorbox{promptbox}[2][]{%
    colback=gray!5, colframe=gray!40,
    colbacktitle=gray!20, coltitle=black,
    fonttitle=\bfseries, title={#2},
    arc=1mm, boxrule=0.5pt, #1}
\newcommand{\hallucination}[1]{\colorbox{red!15}{\textcolor{darkred}{#1}}}

\title{HALT: Hallucination Assessment via Log-probs as Time series}

\author{
 \textbf{Ahmad Shapiro}$^{1,2}$\thanks{\ Work done prior to joining Microsoft.\\\textbf{Correspondence:} \href{mailto:ahmad.shapiro@gatech.edu}{ahmad.shapiro@gatech.edu}},
  \textbf{Karan Taneja}$^{1}$,
  \textbf{Ashok Goel}$^{1}$
\\
\\
 $^1$Georgia Institute of Technology \quad $^2$Microsoft
\\
 \small{
   \textbf{Project Page:} \href{https://halt-hallucination-detection.github.io/halt-v1}{https://halt-hallucination-detection.github.io}
 }
}
\begin{document}
\maketitle
\begin{abstract}
Detecting hallucinations in Large Language Model (LLM) outputs is critical for safe deployment, yet existing methods require white-box access to internals, depend on multiple generations or auxiliary judge models, or collapse token-level uncertainty into scalar summaries that discard temporal structure. We introduce \textbf{HALT}, a lightweight grey-box detector that treats the top-20 log-probabilities of an LLM generation as a multivariate time series and processes them with a small bidirectional GRU (5M parameters), in a single forward pass and using no model internals. We evaluate on \textbf{HUB}, a unified protocol reorganizing existing hallucination datasets into ten capability clusters spanning factual and logical failure modes. HALT matches or surpasses Lettuce, a fine-tuned ModernBERT-base detector, on macro-$F_1$ while using $\sim$$30\times$ fewer parameters; a controlled comparison against a text classifier trained on HALT's own data confirms this advantage comes from the log-probability signal itself rather than from differences in training data, and the same recipe applies across eight LLM families from 360M to 70B parameters.
\end{abstract}

\section{Introduction}
\label{sec:intro}

Large Language Models (LLMs) have achieved remarkable progress in generating fluent and coherent text, yet they remain prone to \emph{hallucinations}: outputs that are verifiably false or unsupported by evidence. \citet{Li2023HaluEval} report that GPT-3.5 introduces unverifiable details in nearly 19.5\% of user queries, and similar failure rates have been documented across model families and tasks \citep{Ji2023Survey}. As LLMs are deployed in increasingly safety-critical settings, reliable post-hoc detection of hallucinations has become essential.

Existing approaches fall into three camps, each with deployment costs. \textbf{White-box} methods inspect internal states such as hidden activations or attention maps \citep{llmcheck,inside,Azaria2023Lying}, requiring model access and per-dataset layer search or training. \textbf{Black-box and multi-sample} methods rely on multiple generations \citep{Manakul2023SelfCheckGPT}, auxiliary judges \citep{Lan2024CriticBench}, or external retrieval \citep{Mishra2024FAVA,ragbench}, introducing latency or monetary cost. \textbf{Aggregate-statistics} methods compress token-level uncertainty into scalars such as perplexity \citep{Varshney2023LowConf,Quevedo2024TokenProb}, which is cheap but discards temporal information about \emph{how} uncertainty evolves.

\begin{figure}[t]
\centering
\resizebox{0.8\linewidth}{!}{
\begin{tikzpicture}[
    node distance=0.7cm,
    auto,
    box/.style={draw, fill=white, rounded corners, minimum height=2.2em, minimum width=4.5cm, align=center, drop shadow, font=\small},
    llm/.style={box, fill=orange!10, draw=orange!50, thick},
    data/.style={box, fill=blue!8, draw=blue!40, thick, font=\small\ttfamily},
    model/.style={box, fill=purple!8, draw=purple!50, thick},
    outbox/.style={box, fill=red!8, draw=red!50, thick},
    arrow/.style={-Stealth, thick, gray!80}
]
    \node[font=\small] (input) {User Prompt};
    \node[llm, below=of input] (llm) {\textbf{Target LLM} (Black Box)};
    \node[data, below=of llm] (logprobs) {$\bm{\ell}_{1:T} \in \mathbb{R}^{T \times 20}$};
    \node[box, fill=teal!8, draw=teal!50, thick, below=of logprobs, font=\footnotesize, align=center] (features) {Feature Extraction\\\scriptsize (AvgLogP, $H_{\text{overall}}$, $H_{\text{alts}}$, RankProxy, $\Delta H_{\text{dec}}$)};
    \node[data, below=of features] (concat) {$\tilde{\bm{\ell}}_{1:T} \in \mathbb{R}^{T \times 25}$};
    \node[model, below=of concat] (halt) {\textbf{HALT}: Bi-GRU + Top-$q$ Pool (5M params)};
    \node[outbox, below=of halt] (output) {$P(\text{hallucinated}) \in [0,1]$};
    
    \draw[arrow] (input) -- (llm);
    \draw[arrow] (llm) -- (logprobs);
    \draw[arrow] (logprobs) -- (features);
    \draw[arrow] (features) -- (concat);
    \draw[arrow] (concat) -- (halt);
    \draw[arrow] (halt) -- (output);
    
    % Bypass arrow showing raw log-probs also feed into concatenation
    \draw[arrow, rounded corners] (logprobs.east) -- ++(1.2,0) |- node[pos=0.25, right, font=\scriptsize, align=left, xshift=2pt] {raw $\bm{\ell}_t$} (concat.east);
\end{tikzpicture}
}
\caption{\textbf{HALT architecture.} Top-20 log-probabilities from the target LLM are augmented with five engineered features and processed by a 5M-parameter Bi-GRU with Top-$q$ pooling to produce a hallucination probability.}
\label{fig:halt_arch}
\end{figure}

This leaves an underexplored point in the design space: \emph{the sequence of top-$K$ log-probabilities returned by most modern APIs and serving frameworks} (e.g., vLLM, \citealp{vllm}). Top-$K$ log-probs (typically $K \le 20$) are richer than a single scalar but far cheaper than full distributions or repeated generations, and require no access to model internals. We argue that the \emph{temporal shape} of this signal carries information about hallucinations that aggregate statistics discard.

We introduce \textbf{HALT} (\underline{H}allucination \underline{A}ssessment via \underline{L}og-probs as \underline{T}ime series), a lightweight grey-box detector that treats the top-20 log-probability vectors of an LLM generation as a multivariate time series and processes them with a small bidirectional GRU \citep{gru}. HALT relies on only a single forward pass, no model internals, and no auxiliary models. We evaluate two trained variants---HALT-L and HALT-Q, on Llama 3.1-8B and Qwen 2.5-7B respectively---and show that the same architecture and training procedure transfer across eight LLM families from 360M to 70B parameters, with a separately trained detector per LLM. To enable broad evaluation, we assemble \textbf{HUB} (Hallucination detection Unified Benchmark), which reorganizes existing labeled data (FAVA, RAGTruth, HaluEval, CriticBench) into ten LLM capability clusters spanning factual and logical (reasoning-induced) failure modes under a unified evaluation protocol. HUB introduces no new annotations.

On HUB, HALT matches or surpasses Lettuce \citep{LettuceDetect}, a fine-tuned ModernBERT-base detector, on macro-$F_1$ while using approximately $30\times$ fewer parameters. The advantage is most pronounced on reasoning-heavy clusters where surface-form cues are scarce; Lettuce remains stronger on RAGTruth, where it is trained in-domain. HALT also remains competitive against multi-sample (SelfCheckGPT, \citealp{Manakul2023SelfCheckGPT}) and high-$K$ (LOS-Net, \citealp{barshalom2025losnet}) methods while operating in a strictly cheaper regime.

\paragraph{Contributions.}
\textbf{HALT}, a 5M-parameter bidirectional GRU to detect hallucinations from top-20 log-probability sequences. HALT operates in a single forward pass with no access to internals, surface text, or external references, and we release trained variants for two LLM families.
\textbf{HUB}, a unified evaluation protocol that organizes existing hallucination datasets (FAVA, RAGTruth, HaluEval, CriticBench) into ten capability clusters spanning factual and logical hallucinations, enabling consistent cross-capability comparison of detection methods. \textbf{A controlled empirical comparison} against single-pass alternatives \citep{Manakul2023SelfCheckGPT,barshalom2025losnet} showing that low-$K$ temporal modeling can match or exceed methods using multiple generations or much larger $K$. \textbf{Architecture, feature, and top-$K$ ablations }isolating which design choices drive performance, together with cross-scale evaluation across eight LLM families (360M--70B parameters).

\section{HUB: A Unified Benchmark}
\label{sec:hub}

Existing hallucination detection benchmarks cover narrow slices of LLM capability: RAGTruth \citep{Niu2024RAGTruth} targets three reference-based tasks (Data-to-Text, QA, Summarization), HaluEval \citep{Li2023HaluEval} covers QA, Dialogue, and Summarization, and FAVA \citep{Mishra2024FAVA} focuses on knowledge-intensive queries. None systematically include reasoning tasks, where LLMs increasingly fail despite producing fluent output. We assemble HUB to address this gap not by collecting new annotations but by reorganizing existing labeled data into a unified evaluation protocol.

\paragraph{Factual versus logical hallucinations.}

Following the faithfulness--factuality distinction of \citet{Ji2023Survey}, we treat a hallucination as any output unfaithful to its task specification. \emph{Factual hallucinations} violate factuality: they contradict established reality even when correct context is provided, as in RAG settings where the model ignores grounding. \emph{Logical hallucinations} are a faithfulness failure: the model correctly identifies the premises but produces inferences unfaithful to the task's logical constraints---arithmetic mistakes, invalid steps, mis-classifications, or inconsistent chains. Figure~\ref{fig:logical_hallucination_example} shows a logical hallucination where the model both misclassifies a non-fruit (\emph{carrot}) and errs in the summation. In \citet{Ji2023Survey}'s framework, faithfulness means the output does not introduce content unsupported by, or inconsistent with, the given input---typically applied to source-grounded settings where an unfaithful output contradicts retrieved context. We extend this to a different kind of ``given'': in a math or logic problem, the premises and the task's inferential rules play the same role retrieved context plays in RAG, and an invalid inference step is a conclusion not entailed by those premises---the same structural failure, applied to task constraints rather than a document.

\begin{figure}[!t]
\centering
\resizebox{\linewidth}{!}{
\begin{minipage}{\linewidth}
    \begin{promptbox}{User Prompt}
    I have a peach, a strawberry, a banana, a carrot, two apples, and four grapes. How many fruits do I have?
    \end{promptbox}
    
    % \vspace{0.05cm}
    
    \begin{promptbox}[colback=red!3, colframe=red!30]{Model Response}
    Let's think step by step. I will identify the fruits and their quantities:
    peach (1), strawberry (1), banana (1), \hallucination{carrot (1)}, apples (2) and grapes (4).
    Adding the quantities: \hallucination{$1 + 1 + 1 + 1 + 2 + 4 = 11$}.
    
    So the answer is \hallucination{11}.
    \end{promptbox}
\end{minipage}
}
\caption{\textbf{Example of a logical hallucination.} The model commits two distinct reasoning errors: (i) a \textit{classification error}, incorrectly including a carrot (a vegetable) in the fruit count, and (ii) an \textit{arithmetic error}, reporting the sum $1{+}1{+}1{+}1{+}2{+}4$ as 11 rather than the correct value 10. The correct answer is 9 (the carrot should be excluded). Both errors are highlighted in red. The sample originates from the \emph{ObjectCounting} shard of CriticBench \citep{Lan2024CriticBench} in the Algorithmic Reasoning cluster.}
\label{fig:logical_hallucination_example}
\end{figure}

\paragraph{Composition.}
HUB organizes existing datasets into ten capability clusters: five \emph{reasoning-oriented} (Algorithmic, Commonsense, Mathematical, Symbolic, Code Generation) and five \emph{general-purpose} (Chat, Data-to-Text, Question Answering, Summarization, World Knowledge). Reasoning clusters are built from CriticBench \citep{Lan2024CriticBench}, whose annotations cover both reasoning-trace errors and final-answer errors. General-purpose clusters draw from FAVA \citep{Mishra2024FAVA}, RAGTruth \citep{Niu2024RAGTruth}, and HaluEval \citep{Li2023HaluEval}. Each cluster aggregates multiple source datasets where available, supporting cross-dataset generalization within a capability. Table~\ref{tab:hub_stats} reports cluster-level statistics; per-dataset details appear in Appendix~\ref{app:hub_details}.

\paragraph{Splits and generalization protocol.}
To stress-test generalization, training is restricted to four general-purpose clusters (Chat, Data-to-Text, Summarization, QA), providing approximately 60K labeled responses while leaving six clusters---all five reasoning clusters plus World Knowledge---held out from training. We avoid the term ``zero-shot'' here: model selection (checkpointing, hyperparameters, thresholds) uses a pooled validation metric spanning every cluster, including the six held-out ones, so training and selection exposure are not identical---though a weaker form of leakage than training on them directly, and mitigated by construction, since validation and test within each held-out cluster draw from disjoint source datasets rather than disjoint examples of one dataset (e.g.\ for Mathematical, validation is TabMWP while test is GSM8K, MATH, and AQuA; full correspondences in Appendix~\ref{app:taxonomy_alignment}). A single global threshold is tuned once on the pooled validation set and applied uniformly, not per-cluster. For FAVA Annotations and RAGTruth we additionally retain the original test splits for direct comparison with prior baselines.

Hallucination ratios vary sharply across clusters, from $\sim$40\% (Chat, Summarization) to $\sim$95\% (World Knowledge). This motivates \textbf{macro-$F_1$} as our primary metric, detailed in Section~\ref{sec:setup}.

\paragraph{Annotation reliability.}
HUB introduces no new annotations. Labels are inherited from the source datasets: FAVA, RAGTruth, and HaluEval contribute human and crowd-sourced annotations, and CriticBench contributes a hybrid pipeline combining rule-based heuristics, GPT-4 judging, and human adjudication on disagreements. To verify that CriticBench's failure-mode taxonomy aligns with our definition of logical hallucination, we manually inspected samples from each reasoning cluster and found approximately 90\% agreement with the original labels. We discuss annotation provenance as a known limitation in Section~\ref{sec:limitations} and inherit the licensing terms of the source datasets.

\section{Method}
\label{sec:method}

\subsection{Setup}
Let $\mathcal{M}_\theta$ be an autoregressive LLM. At each generation step $t$, $\mathcal{M}_\theta$ produces a distribution over its vocabulary, of which we observe only the top-20 log-probabilities:
\begin{equation}
\bm{\ell}_t = \bigl(\log p_t^{(0)}, \log p_t^{(1)}, \ldots, \log p_t^{(19)}\bigr) \in \mathbb{R}^{20},
\end{equation}
where $\ell_t^{(0)}$ is the \emph{selected} token (not necessarily greedy under sampling) and indices $1$--$19$ are the top-19 alternatives sorted by probability in descending order, regardless of the selected token's rank (used in the $\mathrm{RankProxy}$ definition, Section~\ref{sec:features}). The serving framework always returns the selected token's log-probability; when it falls outside the top-19, we take the union of both sets. A response of length $T$ yields $\bm{\ell}_{1:T} \in \mathbb{R}^{T \times 20}$, obtained by teacher-forcing the labeled response into $\mathcal{M}_\theta$ via vLLM \citep{vllm}. The choice $K=20$ is justified in Section~\ref{sec:topk}.

\subsection{Token-Level Features}
\label{sec:features}
For each timestep, we augment $\bm{\ell}_t$ with five engineered features capturing local calibration behavior:

\begin{itemize}
    \item \textbf{Average log-probability} ($\mathrm{AvgLogP}$) -- mean over the top-20 entries; surrogate for distributional sharpness.
    \item \textbf{Rank proxy} ($\mathrm{RankProxy}$) -- the position of the selected token among the top-20 alternatives, indicating non-greedy selections.
    \item \textbf{Top-$K$ entropy} ($H_{\mathrm{overall}}$) -- entropy of the renormalized top-20 distribution; high values indicate indecision among the leading candidates.
    \item \textbf{Alternatives entropy} ($H_{\mathrm{alts}}$) -- entropy over the top-20 distribution \emph{excluding} the selected token, isolating disagreement among competitors.
    \item \textbf{Decision-entropy delta} ($\Delta H_{\mathrm{dec}}$) -- temporal change in binary entropy between the selected token and its strongest alternative; flags rapid certainty--uncertainty transitions.
\end{itemize}

These five were not selected by exhaustive search; they span four complementary families: (i) \emph{magnitude} ($\mathrm{AvgLogP}$), (ii) \emph{rank} of the selected token among alternatives ($\mathrm{RankProxy}$), (iii) \emph{distributional-shape entropy} over two supports---$H_{\mathrm{overall}}$ (full top-$K$) and $H_{\mathrm{alts}}$ (alternatives only, isolating competitor disagreement from the selected token's own confidence)---and (iv) \emph{temporal dynamics} ($\Delta H_{\mathrm{dec}}$), capturing change rather than level. Each targets a distinct aspect of token-level uncertainty, consistent with the largely non-overlapping, additive gains in the incremental ablation (Table~\ref{tab:feat_ablation}).

Concatenating these with the raw $\bm{\ell}_t$ yields a 25-dimensional feature vector per timestep:
\begin{equation}
\tilde{\bm{\ell}}_t = \bigl[\phi(\bm{\ell}_t) \,\big|\, \bm{\ell}_t\bigr] \in \mathbb{R}^{25}.
\end{equation}
Full derivations and the truncated-softmax normalization used for $H_{\mathrm{overall}}$ and $H_{\mathrm{alts}}$ are provided in Appendix~\ref{app:features_extended}.

\subsection{Architecture}
\label{sec:architecture}
Figure~\ref{fig:halt_arch} summarizes the architecture. The 25-dimensional feature sequence $\tilde{\bm{\ell}}_{1:T}$ is first normalized (LayerNorm) and projected to a 128-dimensional embedding via a two-layer GELU MLP. The projected sequence is encoded by a bidirectional GRU \citep{gru} with 5 layers and hidden size 256 (dropout 0.4 between recurrent layers). Variable-length responses are handled with packed sequences and a boolean attention mask, ensuring padded positions do not influence hidden dynamics.

We aggregate the bidirectional GRU outputs ($H \in \mathbb{R}^{T \times 512}$) using \textbf{Top-$q$ pooling}: timesteps are scored by their hidden-state $\ell_2$ norm, padding positions are masked out, and the top $q = 15\%$ of timesteps are averaged. This concentrates the classifier on the most informative regions of the sequence (e.g., bursts of uncertainty around hallucination onsets) rather than diluting the signal across all tokens. We compare Top-$q$ pooling against mean, max, and attention-based pooling in Appendix~\ref{app:pooling_ablation}.

The pooled vector passes through a linear head to a single logit, trained with \texttt{BCEWithLogitsLoss}. The full model has $\sim$5M parameters; training details are in Appendix~\ref{app:training}.

\subsection{Why a Temporal Model? (Architecture Ablation)}
\label{sec:arch_ablation}

A natural question is whether the temporal structure of $\tilde{\bm{\ell}}_{1:T}$ is necessary at all, or whether aggregate statistics over the sequence would suffice. We compare HALT's Bi-GRU encoder against both non-temporal baselines and alternative temporal architectures using identical features and training data. Table~\ref{tab:arch_ablation} reports macro-$F_1$ averaged over HUB clusters and overall.

We add two attention-based architectures: a parameter-matched Transformer encoder (self-attention over $\tilde{\bm{\ell}}_{1:T}$, same features, Top-$q$ pooling, and training data, 5.47M params vs.\ HALT-L's 5.34M) and a parameter-matched reimplementation of LOS-Net \citep{barshalom2025losnet} (Transformer with rank encoding and CLS pooling, faithful to its published design) evaluated under HUB's own splits and $K=20$ rather than its original top-50/top-1000 regime, both retuned with standard Transformer defaults after an initial run with HALT-L's Bi-GRU-tuned optimizer settings underperformed.

\begin{table}[t]
\centering
\small
\caption{Architecture comparison on HUB using Llama 3.1-8B log-probabilities. Non-temporal models operate on per-sequence aggregate statistics; temporal models operate on the full $\tilde{\bm{\ell}}_{1:T}$ sequence. LOS-Net-faithful and Transformer-encoder are parameter-matched to HALT-L ($\sim$5.4M params) and trained under identical data and $K{=}20$ with tuned hyperparameters.}
\label{tab:arch_ablation}
\begin{tabular}{llcc}
\toprule
\textbf{Type} & \textbf{Model} & \textbf{Avg.\ F1} & \textbf{Overall F1} \\
\midrule
Non-temporal & LR & 0.385 & 0.430 \\
Non-temporal & XGBoost             & 0.372 & 0.418 \\
\midrule
Temporal     & 1D-CNN              & 0.510 & 0.565 \\
Temporal     & Bi-RNN              & 0.507 & 0.551 \\
Temporal     & Uni-GRU             & 0.543 & 0.600 \\
Temporal     & LOS-Net    & 0.540 & 0.613    \\
Temporal     & Transformer & 0.560 & 0.618    \\
Temporal     & Bi-LSTM             & 0.591 & 0.655 \\
Temporal     & \textbf{Bi-GRU} & \textbf{0.630} & \textbf{0.670} \\
\bottomrule
\end{tabular}
\end{table}

Three patterns emerge. Temporal modeling provides a substantial gain: the best non-temporal baseline reaches only 0.430 Overall $F_1$, while every temporal architecture exceeds 0.550. This 12--24 point gap empirically refutes the implicit assumption of aggregate-statistics methods \citep{Varshney2023LowConf,Quevedo2024TokenProb} that collapsing the log-probability sequence loses no useful information. Gating matters (Bi-RNN trails Bi-LSTM and Bi-GRU by 5--10 points), and bidirectionality contributes another 4--7 points, consistent with the intuition that hallucination signatures often become recognizable in retrospect. Under an identical parameter budget, data, and $K$, both attention-based alternatives trail Bi-GRU by 7--9 Avg.\ $F_1$ points (0.540 and 0.560 vs.\ 0.630); retuning narrowed but did not close this gap, consistent with a genuine generalization ceiling on HUB's training scale rather than an optimization artifact.

\subsection{What Signal Do the Features Carry?}
\label{sec:feat_ablation}

Within HALT's input, we further isolate the contribution of each engineered feature by training incremental variants that begin from raw log-probabilities and progressively add features (Table~\ref{tab:feat_ablation}). We also include a variant that drops the raw log-probability channels and retains only the five engineered features, still processed by the full Bi-GRU and Top-$q$ pooling---this isolates the marginal contribution of the raw channels given a temporal encoder already in place, and should not be read as a non-temporal baseline (that comparison is Table~\ref{tab:arch_ablation}, Section~\ref{sec:arch_ablation}).

\begin{table}[t]
\centering
\small
\caption{Feature ablation on HUB. Incremental rows add one engineered feature at a time on top of raw top-20 log-probabilities. The last row drops the raw log-probability channels but keeps the full temporal encoder; it isolates the raw channels' marginal contribution, not the contribution of temporal modeling (see Table~\ref{tab:arch_ablation} for that comparison).}
\label{tab:feat_ablation}
\begin{tabular}{lcc}
\toprule
\textbf{Variant} & \textbf{Avg.\ F1} & \textbf{Overall F1} \\
\midrule
Raw log-probs only          & 0.557 & 0.616 \\
\quad + $\mathrm{AvgLogP}$         & 0.562 & 0.618 \\
\quad + $H_{\mathrm{alts}}$       & 0.575 & 0.647 \\
\quad + $\mathrm{RankProxy}$        & 0.600 & 0.657 \\
\quad + $H_{\mathrm{overall}}$    & 0.614 & 0.660 \\
\textbf{Full (+ $\Delta H_{\mathrm{dec}}$)} & \textbf{0.630} & \textbf{0.670} \\
\midrule
Full, no raw log-probs (temporal) & 0.567 & 0.638 \\
\bottomrule
\end{tabular}
\end{table}

Raw top-20 log-probabilities alone already provide a strong signal (0.616 Overall $F_1$), exceeding the best non-temporal baseline by 19 points and confirming that the bulk of the signal lies in the \emph{shape} of the top-$K$ distribution. Engineered features contribute an additional $\sim$5 points, with the largest gains from $H_{\mathrm{alts}}$ ($+3.1$) and $\mathrm{RankProxy}$ ($+1.0$). Dropping the raw log-probability channels while keeping the full temporal encoder costs 3.2 Overall $F_1$ points, showing the raw channels and the engineered features are complementary rather than redundant; this is a separate quantity from the 12--24 point non-temporal-vs-temporal gap in Table~\ref{tab:arch_ablation}, not a competing estimate of it. Per-feature attribution is in Appendix~\ref{app:feature_attribution}.

\paragraph{Ruling out a length confound.} A natural alternative explanation is that HALT is really picking up on response length, which correlates with both uncertainty accumulation and task difficulty. Predicting the hallucination label from length alone gives an AUROC of 0.56 (barely above chance) on the HUB test set, vs.\ 0.72 from HALT-L's predicted probability, and the two classes separate along the probability axis at every length scale examined---evidence against length as the dominant driver.

\subsection{Why \texorpdfstring{$K=20$}{K=20}?}
\label{sec:topk}

The choice of $K$ determines how much of the predictive distribution HALT observes. Increasing $K$ provides richer temporal signals but also increases extraction cost and storage. We study this trade-off in two ways: by varying $K$ during HALT-L training, and by measuring the cumulative probability mass captured by top-$K$ across diverse LLMs.

\begin{table}[t]
\centering
\small
\caption{Effect of $K$ on HALT-L (LLaMA 3.1-8B log-probabilities). Performance saturates near $K=20$.}
\label{tab:topk}
\begin{tabular}{cccc}
\toprule
\textbf{$K$} & \textbf{Avg.\ F1} & \textbf{Overall F1} & \textbf{Top-$K$ mass (12 LLMs)} \\
\midrule
1   & 0.546 & 0.593 & 57--65\% \\
5   & 0.556 & 0.635 & 88--98\% \\
10  & 0.604 & 0.658 & 93--99\% \\
15  & 0.612 & 0.661 & 94--99\% \\
\textbf{20}  & \textbf{0.630} & \textbf{0.670} & \textbf{95--99.7\%} \\
\bottomrule
\end{tabular}
\end{table}

Performance increases monotonically with $K$ from 0.593 to 0.670, tapering off beyond $K{=}10$. Top-20 captures 95--99.7\% of the predictive mass across 12 LLMs, so entropy- and rank-based features computed at $K{=}20$ closely approximate their full-vocabulary counterparts; beyond this point additional log-probabilities contribute negligible mass at increased serving cost. We adopt $K{=}20$ as the principled operating point, which also matches the limits exposed by most APIs.

\section{Experiments}
\label{sec:experiments}

\subsection{Setup}
\label{sec:setup}

We evaluate HALT on the HUB benchmark (Section~\ref{sec:hub}) and on the FAVA Annotations and RAGTruth subsets retained for comparison with prior published baselines. We compare against four families of methods: (i) \emph{white-box} detectors that inspect internal model states \citep{llmcheck}, (ii) \emph{aggregate-statistics} baselines that reduce token-level uncertainty to scalar summaries (perplexity, $H_{\mathrm{overall}}$, $H_{\mathrm{alts}}$, $\Delta H_{\mathrm{dec}}$), (iii) the \emph{span-based} Lettuce detector \citep{LettuceDetect}, a fine-tuned ModernBERT-base model trained to predict hallucinated spans, and (iv) \emph{single-pass alternatives} including SelfCheckGPT \citep{Manakul2023SelfCheckGPT} and LOS-Net \citep{barshalom2025losnet}.

Aggregate-statistics baselines reduce the per-token sequence to a scalar (mean, or maximum for $\mathrm{RankProxy}$); Lettuce span predictions are converted to a sentence-level label by flagging any span with probability $\geq 0.5$. All thresholds are tuned on the HUB validation set and fixed for test.

Main-results tables report the mean over three seeds; per-cluster 95\% CIs are $\pm$0.04--0.07 macro-$F_1$ on larger clusters and $\pm$0.05--0.12 on the smallest held-out clusters, where rankings should be read with corresponding caution.

\paragraph{Why macro-$F_1$.}
HUB clusters exhibit hallucination ratios from $\sim$30\% to $\sim$95\%. Under this skew, micro-averaging is dominated by the large general-purpose clusters and masks failures on smaller, more imbalanced reasoning clusters where detection matters most; a detector predicting the majority class on World Knowledge (80\% hallucinated) achieves high accuracy while contributing nothing. Macro-$F_1$ weights each cluster equally and penalizes errors from both classes, making it the appropriate primary metric for cross-capability generalization. Throughout, \emph{per-cluster} scores are macro-$F_1$ within a cluster, \emph{Overall} is macro-$F_1$ over the pooled test set, and \emph{Avg.} is the unweighted mean of per-cluster scores.

\paragraph{Evaluation regime.}
HUB results (Tables~\ref{tab:hub_main}--\ref{tab:fava_ragtruth}) use \emph{teacher-forced} log-probabilities: each labeled response is fed through the target LLM and its top-20 log-probabilities recorded. Because HUB's responses originate from heterogeneous sources, teacher-forcing is what makes a controlled, comparable log-probability signal obtainable across all clusters under a single target LLM. HALT thus detects whether a response is inconsistent with the target LLM's distribution over faithful text---a detection target directly relevant to the common setting of scoring outputs from a model one does not control. The single-pass alternatives in Section~\ref{sec:single_pass} instead use each LLM's own free-running generations; that HALT remains effective there provides evidence that the signal is not an artifact of the teacher-forced regime. We discuss the teacher-forced/self-generated distinction further in Limitations.

\subsection{Main Results on HUB}
\label{sec:hub_results}

Table~\ref{tab:hub_main} reports macro-$F_1$ across HUB's ten capability clusters. HALT-L (trained on Llama 3.1-8B log-probabilities) achieves the highest overall macro-$F_1$ (67.0) and average-across-clusters macro-$F_1$ (63.0), winning 7 of 10 clusters. HALT-Q (trained on Qwen 2.5-7B log-probabilities) is consistently second across most clusters; it received only a light hyperparameter sweep ($+3$ Overall $F_1$ over direct transfer), a likely conservative estimate of its potential.

\begin{table*}[t]
\centering
\small
\caption{Macro-$F_1$ on HUB test clusters (teacher-forced log-probabilities). Aggregate baselines use Llama 3.1-8B; \textbf{bold}/\underline{underline} mark best/second-best per cluster. HALT's training covers the capabilities Lettuce and Text-ModernBERT were trained on (Data-to-Text, QA, Summarization, Chat); all five reasoning clusters (including Code Generation) and World Knowledge are held out from training for all methods. Text-ModernBERT is a ModernBERT-base classifier fine-tuned end-to-end on HUB's own training split---same architecture family as Lettuce, but trained on HALT's own training distribution rather than RAGTruth.}
\label{tab:hub_main}
\resizebox{\textwidth}{!}{
\begin{tabular}{l|cccc|cc|cc}
\toprule
\textbf{Cluster} & \textbf{PPL} & \textbf{$H_{\mathrm{overall}}$} & \textbf{$\Delta H_{\mathrm{dec}}$} & \textbf{$H_{\mathrm{alts}}$} & \textbf{Lettuce} & \textbf{Text-ModernBERT} & \textbf{HALT-L} & \textbf{HALT-Q} \\
\midrule
Algorithmic        & 24.2 & 24.2 & 26.4 & 24.5 & 24.2 & \underline{43.1} & \textbf{76.8} & 32.7 \\
Chat               & 35.1 & 34.6 & 37.4 & 36.5 & 41.5 & \underline{60.0} & \textbf{60.2} & 58.6 \\
Code Generation    & 43.1 & 38.1 & \underline{62.0} & \textbf{66.7} & 38.1 & 46.2 & 47.7 & 39.7 \\
Commonsense        & 33.3 & 32.0 & 35.2 & 34.3 & 41.1 & \underline{53.4} & \textbf{56.7} & 41.3 \\
Data-to-Text       & 39.2 & 39.2 & 42.2 & 39.2 & \textbf{83.4} & 60.1 & 72.9 & \underline{73.0} \\
Mathematical       & 44.1 & 42.0 & \underline{67.0} & 61.6 & 41.8 & 54.1 & \textbf{72.7} & 62.9 \\
QA                 & 28.3 & 25.7 & 49.4 & 43.4 & \textbf{77.3} & 71.0 & \underline{74.1} & 68.8 \\
Summarization      & 24.5 & 24.5 & 45.0 & 33.0 & 59.7 & 59.6 & \underline{66.9} & \textbf{70.8} \\
Symbolic           & 24.6 & 24.6 & 24.9 & 24.6 & 33.4 & \underline{58.6} & \textbf{65.4} & 49.8 \\
World Knowledge    & 44.5 & 44.5 & 44.5 & 44.5 & 44.5 & 52.8 & \textbf{76.9} & \underline{58.5} \\
\midrule
Overall            & 34.0 & 32.9 & 48.0 & 42.9 & \underline{64.0} & 61.9 & \textbf{67.0} & 62.7 \\
Avg.\ across clusters & 34.1 & 32.9 & 43.4 & 40.8 & 48.5 & \underline{55.9} & \textbf{63.0} & 55.6 \\
\bottomrule
\end{tabular}}
\end{table*}

HALT's wins are concentrated on the five reasoning clusters (Algorithmic, Symbolic, Mathematical, Commonsense, World Knowledge: $+15$ to $+53$ over Lettuce), held out from training for \emph{both} methods---a symmetric comparison, not an in-domain advantage. On the general-purpose clusters the comparison is fair in the other direction: RAGTruth, Lettuce's training set, is a subset of HALT's training data, so both are in-domain on Data-to-Text, QA, and Summarization, where Lettuce benefits from ModernBERT's pretraining on trillions of text tokens against HALT's random initialization on log-probabilities alone. On the reasoning clusters the aggregate and span baselines collapse to near-constant-class prediction (e.g., five methods at 44.5 on World Knowledge): the comparison there is categorical rather than graded---HALT extracts usable signal where the others fail entirely.

\paragraph{Isolating modality from training distribution.} The comparison against Lettuce leaves one variable confounded: modality (text vs.\ log-probabilities) \emph{and} training distribution (RAGTruth vs.\ HUB). To isolate modality alone, we add \textbf{Text-ModernBERT}, a ModernBERT-base classifier fine-tuned end-to-end on HUB's own training split---the same four clusters HALT trains on. It is closely matched to HALT-L on Chat (60.0 vs.\ 60.2) but trails on Data-to-Text, QA, and Summarization (Table~\ref{tab:hub_main}), and the gap widens further on every held-out cluster (+1.5 to +33.7 macro-$F_1$). With modality now the only variable, the log-probability signal's advantage over surface text holds at every capability level and is largest on the clusters neither model trained on.

\subsection{Comparison on FAVA and RAGTruth}
\label{sec:fava_ragtruth}

We additionally report results on the FAVA Annotations and RAGTruth test sets in isolation (Table~\ref{tab:fava_ragtruth}).

\begin{table}[t]
\centering
\small
\caption{Performance on FAVA Annotations and RAGTruth test sets. Macro-$F_1$ is omitted (--) where not reported by prior baselines. Lettuce is trained on RAGTruth; the FAVA Model is a 7B fine-tuned Llama paired with a retriever (approximately 1{,}400$\times$ HALT's parameter count).}
\label{tab:fava_ragtruth}
\resizebox{\columnwidth}{!}{
\begin{tabular}{l|ccc|ccc}
\toprule
\multirow{2}{*}{\textbf{Method}} 
& \multicolumn{3}{c|}{\textbf{FAVA Annotations}} 
& \multicolumn{3}{c}{\textbf{RAGTruth}} \\ 
& AUROC & $F_1$ & Mac-$F_1$ & AUROC & $F_1$ & Mac-$F_1$ \\
\midrule
\multicolumn{7}{c}{\textit{White-Box}} \\
LLM-Check (Attn)    & \textbf{68.2} & 70.5 & -- & 58.3 & 57.2 & -- \\
\midrule
\multicolumn{7}{c}{\textit{Aggregate Statistics}} \\
$H_{\mathrm{overall}}$ & 53.9 & 80.0 & 40.3 & 37.0 & 51.8 & 25.9 \\
$H_{\mathrm{alts}}$    & 48.9 & 80.0 & 44.0 & 66.3 & 54.0 & 37.2 \\
$\Delta H_{\mathrm{dec}}$ & 50.9 & 80.0 & \underline{44.2} & 65.7 & 56.2 & 47.3 \\
PPL                    & 46.1 & 80.0 & 41.8 & 63.0 & 51.8 & 26.0 \\
\midrule
\multicolumn{7}{c}{\textit{Black/Grey-Box}} \\
FAVA Model (7B)        & 53.3 & \underline{79.9} & -- & -- & -- & -- \\
Lettuce                & 45.8 & \textbf{80.7} & 40.3 & \textbf{82.6} & \textbf{74.5} & \textbf{80.0} \\
LLM-Check (Hidden)  & 57.1 & 65.4 & -- & 57.2 & 47.5 & -- \\
\textbf{HALT-L (ours)} & \underline{61.3} & 77.9 & \textbf{61.6} & \underline{70.7} & \underline{59.0} & \underline{65.7} \\
\bottomrule
\end{tabular}}
\end{table}

On FAVA, HALT-L leads macro-$F_1$ (61.6) by 17 points over the next-best aggregate baseline; the class imbalance (67\% positives) collapses Lettuce's vanilla $F_1$ (80.7) to 40.3 macro-$F_1$, reflecting a bias toward predicting hallucinations. HALT-L remains balanced despite being $1{,}400\times$ smaller than FAVA Model and $30\times$ smaller than Lettuce. On RAGTruth, Lettuce wins across all metrics---unsurprising, since it is trained on RAGTruth only---while HALT-L ranks second on every metric without dedicated fine-tuning.

\subsection{Cross-Scale Generalization}
\label{sec:cross_scale}

A natural question is whether HALT's architecture and training procedure depend on a specific LLM size or family. We train independent HALT detectors on log-probabilities from eight LLM families spanning 360M to 70B parameters, using identical architecture and (with one exception noted below) identical hyperparameters. Table~\ref{tab:cross_scale} summarizes the results.

\begin{table}[t]
\centering
\small
\caption{HALT trained on log-probabilities from eight LLMs ranging from 360M to 70B parameters. All models use HALT-L's hyperparameters; only HALT-Q received a light hyperparameter sweep ($+3$ Overall $F_1$).}
\label{tab:cross_scale}
\begin{tabular}{lccc}
\toprule
\textbf{LLM} & \textbf{Params} & \textbf{Overall $F_1$} & \textbf{Avg.\ $F_1$} \\
\midrule
SmolLM     & 360M & 0.593 & 0.527 \\
SmolLM     & 1.7B & 0.609 & 0.539 \\
Llama 3.2 & 3B   & 0.628 & 0.560 \\
HALT-Q (Qwen 2.5)                   & 7B   & 0.627 & 0.556 \\
\textbf{HALT-L (Llama 3.1)}         & \textbf{8B}   & \textbf{0.670} & \textbf{0.630} \\
Qwen 3      & 14B  & 0.625 & 0.526 \\
Qwen 3      & 32B  & 0.641 & 0.555 \\
Llama 3.1 & 70B  & 0.659 & 0.595 \\
\bottomrule
\end{tabular}
\end{table}

HALT achieves stable performance (Overall $F_1$ between 0.59 and 0.67) across an order of magnitude of scale and three architectural families (SmolLM \citep{allal2024SmolLM}, Llama \citep{grattafiori2024llama3herdmodels}, Qwen \citep{yang2025qwen3technicalreport}) using a single hyperparameter set tuned only on HALT-L; the seven other LLMs inherit these directly, a likely conservative estimate given that a brief HALT-Q sweep alone yielded $+3$ Overall $F_1$. The empirical claim is modest and important: \emph{the HALT recipe transfers across scales without per-LLM re-engineering}.

\subsection{Cross-Model Transfer is Limited}
\label{sec:cross_model}

A separate question is whether a HALT detector trained on one LLM's log-probabilities can be applied directly to a different LLM's log-probabilities at inference. We test this by evaluating HALT-L on Qwen 2.5-7B log-probabilities and HALT-Q on Llama 3.1-8B log-probabilities. HALT-L's Overall $F_1$ drops from 0.670 (on Llama log-probs) to 0.540 (on Qwen log-probs), and HALT-Q drops from 0.627 to 0.51. Both remain above the constant-class baseline (0.327) and the weighted-random baseline (0.507) but substantially below in-distribution performance.

This reflects that calibration profiles---entropy ranges, uncertainty distributions, and temporal patterns around hallucination boundaries---are not directly shared across LLM families. In deployment, a separate HALT detector must be trained per target LLM, a one-time cost of minutes of compute on a single GPU; we list this in Limitations (Section~\ref{sec:limitations}).

\subsection{Comparison with Single-Pass Alternatives}
\label{sec:single_pass}

Two recent methods target a similar deployment niche to HALT: SelfCheckGPT \citep{Manakul2023SelfCheckGPT}, based on multi-sample self-consistency, and LOS-Net \citep{barshalom2025losnet}, which uses richer top-$K$ distributions. Unlike all preceding experiments, which use teacher-forced log-probabilities over labeled responses, these two comparisons use each target LLM's (Llama 3.1 8B) own free-running generations, following each method's published evaluation setup.

\begin{table}[t]
\centering
\small
\caption{HALT-L (Llama 3.1-8B, free generation) against a wider set of black-box, white-box, and multi-sample detectors on CoQA, SQuAD, and XSUM (AUROC). \emph{Single gen.}\ marks methods that require only one generation pass. Non-HALT numbers are as reported in \citet{bazarova2026}; HALT-L was not trained on any of these three datasets.}
\label{tab:vs_selfcheckgpt}
\resizebox{\columnwidth}{!}{
\begin{tabular}{l|c|c|ccc}
\toprule
\textbf{Method} & \textbf{Access} & \textbf{Single gen.} & \textbf{CoQA} & \textbf{SQuAD} & \textbf{XSUM} \\
\midrule
SelfCheckGPT-NLI  & Black-box & $\times$ & \textbf{0.91} & 0.65 & \textbf{0.68} \\
Semantic entropy  & Black-box & $\times$ & 0.78 & 0.55 & 0.47 \\
EigenScore        & White-box & $\times$ & 0.80 & 0.45 & 0.50 \\
HaloScope         & White-box & \checkmark & 0.67 & 0.84 & 0.55 \\
LLM-Check         & White-box & \checkmark & 0.54 & 0.49 & 0.52 \\
ReDEEP            & White-box & \checkmark & 0.58 & 0.39 & 0.62 \\
TOHA              & White-box & \checkmark & \underline{0.84} & \underline{0.87} & \underline{0.65} \\
\textbf{HALT-L (ours)} & Grey-box  & \checkmark & 0.65 & \textbf{0.90} & 0.59 \\
\bottomrule
\end{tabular}}
\end{table}

\begin{table}[t]
\centering
\small
\caption{HALT vs.\ LOS-Net (AUROC). HALT-L was not trained on HotPotQA or MOVIES specifically. LOS-Net in-domain is trained on the evaluation dataset using top-50 log-probabilities; LOS-Net cross-domain is trained on MOVIES/IMDB and tested on HotPotQA using top-1000 log-probabilities.}
\label{tab:vs_losnet}
\resizebox{\columnwidth}{!}{
\begin{tabular}{l|ccc}
\toprule
\textbf{Dataset} & \textbf{HALT-L} & \textbf{LOS-Net (in-dom.)} & \textbf{LOS-Net (cross)} \\
                 & (top-20) & (top-50) & (top-1000) \\
\midrule
HotPotQA & 0.65          & \textbf{0.69} & 0.635--0.67 \\
MOVIES   & 0.66          & \textbf{0.77} & 0.67--0.75          \\
\bottomrule
\end{tabular}}
\end{table}

Against SelfCheckGPT-NLI (Table~\ref{tab:vs_selfcheckgpt}), HALT-L outperforms by 25 AUROC points on SQuAD using 20$\times$ fewer generations and no auxiliary NLI model, but loses on XSUM by 9 points---consistent with multi-sample consistency capturing summarization-style hallucinations that single-pass methods miss, while single-pass temporal modeling wins where forced-choice grounding matters more than stylistic consistency. \textit{This is a deliberate cost-accuracy trade-off, not a uniform claim of dominance.} Against the wider set of reference numbers from \citet{bazarova2026}, HALT-L's SQuAD AUROC (0.90) exceeds every method shown, including TOHA (0.87), using only a single forward pass and top-20 log-probabilities; on CoQA and XSUM it is not strongest, consistent with the same trade-off---trading some peak accuracy for requiring neither internals nor resampling.

Against LOS-Net (Table~\ref{tab:vs_losnet}), LOS-Net achieves stronger in-domain performance using top-50 log-probabilities and dataset-specific training, while HALT-L remains within range of LOS-Net's cross-domain numbers on HotPotQA using $50\times$ fewer log-probability values and no dataset-specific training. HALT thus occupies a distinct design point: lower $K$, single-pass, at some cost in peak in-domain accuracy.

\subsection{Efficiency}
\label{sec:efficiency}

A 5M-parameter detector matches or exceeds a 150M-parameter alternative: HALT is approximately $30\times$ smaller than Lettuce, and strictly cheaper than multi-sample methods such as SelfCheckGPT-NLI, which requires 20 additional generations through the target LLM \emph{plus} pairwise inference through an auxiliary BERT-style NLI model. HALT instead operates on log-probabilities emitted as a byproduct of the original generation, paid once per response. We extend discussion about efficiency in Appendix \ref{app:efficiency}.

\section{Related Work}
\label{sec:related_work}

\paragraph{White-box methods.}
A first line of work inspects internal model states: LLM-Check \citep{llmcheck} aggregates attention/hidden-activation statistics, INSIDE \citep{inside} uses the eigenvalue spectrum of hidden-state covariance, ACT-ViT \citep{actvit} probes activation tensors directly, and \citet{Azaria2023Lying} train classifiers on activations to detect statements the model ``knows'' are false. Message-passing over attention graphs \citep{attngraph}, hidden-state trajectory tracking \citep{icrprobe}, and unsupervised internal-state signals \citep{su2024unsupervised} extend this line further. These methods require white-box access and typically per-dataset layer tuning.

\paragraph{Black-box and multi-sample methods.}
A second line operates without internal access by leveraging multiple generations or auxiliary models. SelfCheckGPT \citep{Manakul2023SelfCheckGPT} samples 10--20 responses and flags hallucinations via cross-sample inconsistency; semantic entropy \citep{farquhar2024semantic} clusters generations by meaning-equivalence and uses cluster-level entropy, an idea \citet{vipulanandan2026} extend with a quantum-tensor-network uncertainty estimate at finer granularity, at the cost of the same external clustering model. Fact-level decomposition \citep{factselfcheck}, consistency-under-uncertain-expression checks \citep{consistencyuncertain}, and cross-model verification \citep{verifywhenuncertain} refine self-consistency further. Detection and correction \citep{Mishra2024FAVA} and retrieval-based methods \citep{ragbench,Niu2024RAGTruth} compare against external evidence instead. All introduce additional inference cost, monetary cost, or external dependencies.

\paragraph{Aggregate-statistics methods.}
A third line of work, closer to HALT in input modality, derives scalar uncertainty summaries from the LLM's output distribution: \citet{Varshney2023LowConf} use minimum token-level log-probability as a confidence score, \citet{Quevedo2024TokenProb} aggregate per-token probabilities through hand-crafted reductions, and standard perplexity-based detection averages log-probabilities across the sequence. These are cheap but collapse the response to a single scalar, discarding the temporal structure that Section~\ref{sec:arch_ablation}'s ablation shows carries 12--24 macro-$F_1$ points of additional signal.

\paragraph{LOS-Net.}
Closest to HALT is LOS-Net \citep{barshalom2025losnet}, which also identifies the top-$K$ output distribution as a useful hallucination signal, arrived at independently since LOS-Net predates this paper. LOS-Net differs in three respects: much larger $K$ (top-50 in-domain, top-1000 cross-domain), an attention-based static encoder rather than an explicit temporal sequence, and in-domain dataset-specific training, versus HALT's $K{=}20$, recurrent temporal encoder, and cross-capability evaluation without per-dataset training. The two findings are complementary: both show output distributions carry hallucination signal beyond aggregate statistics, and HALT shows this signal is recoverable at much lower $K$ when temporal structure is modeled explicitly.

\section{Conclusion}
\label{sec:conclusion}

We introduced HALT, a grey-box hallucination detector that treats the top-20 log-probabilities of an LLM generation as a multivariate time series and processes them with a small bidirectional GRU. On HUB---ten capability clusters spanning factual and logical failure modes---HALT matches or surpasses a 30$\times$ larger ModernBERT detector on macro-$F_1$, wins on reasoning clusters where surface-form cues are sparse, and stays competitive against multi-sample and high-$K$ alternatives at lower cost. Architecture and feature ablations isolate the contributions of temporal modeling, and the cross-scale evaluation shows the recipe transfers across eight LLM families.

These results suggest that the temporal shape of top-$K$ log-probabilities is a richer hallucination signal than the aggregate-statistics literature has assumed. Promising directions include a online detection during generation, and span-level localization.

\section*{Limitations}
\label{sec:limitations}

While HALT achieves strong performance across diverse LLMs and capability clusters, several limitations should be acknowledged.

\paragraph{Model-specific detectors.} HALT's signal depends on the calibration profile of a specific LLM. Cross-model transfer experiments (Section~\ref{sec:cross_model}) show substantial performance drops when applying a detector trained on one LLM to log-probabilities from another. In deployment this means a separate detector must be trained per target LLM, a one-time cost on the order of minutes of single-GPU compute, but a real engineering overhead.

\paragraph{Annotation provenance.} HUB inherits labels from its source datasets and introduces no new annotations. FAVA, RAGTruth, and HaluEval contribute human or crowd-sourced labels, while CriticBench labels come from a hybrid pipeline combining rule-based heuristics, GPT-4 judging, and human adjudication on disagreements. Our manual inspection of CriticBench samples found approximately 90\% agreement with the original labels, but the underlying reliance on LLM-assisted labeling for the reasoning clusters remains a constraint on benchmark fidelity.

\paragraph{Validation exposure and small held-out clusters.} Model selection uses a pooled validation metric spanning every cluster, including the six held out from training, so we avoid calling them strictly zero-shot. This is sharpest for World Knowledge: only 12 faithful validation examples, and a hallucination rate shifting from 95\% (validation) to 80\% (test). Since thresholds are tuned globally rather than per cluster (Section~\ref{sec:setup}), these 12 examples do not independently fix its cutoff---but rankings on this cluster should still be read with the wide confidence intervals from Section~\ref{sec:setup} in mind.

\paragraph{English-only evaluation.} All datasets used in HUB are English. HALT's performance on multilingual or code-switched text remains unevaluated, and the temporal calibration patterns we exploit may differ qualitatively in non-English settings, particularly for languages with different tokenization characteristics.

\paragraph{Hyperparameter tuning provenance.} The HALT architecture and training hyperparameters were tuned on HALT-L (Llama 3.1-8B). HALT-Q received a light additional sweep that yielded $+3$ Overall $F_1$; the remaining six LLMs in the cross-scale evaluation use HALT-L's hyperparameters without modification. Their reported numbers are likely conservative estimates of achievable performance.

\paragraph{Teacher-forced evaluation.} HUB results use teacher-forced log-probabilities, whereas deployment scores a model's own free-running generations, where the distribution differs. Forced tokens are not systematically out-of-distribution: mean rank 2.13, top-20 capturing 99.0\% of predictive mass, only 2.07\% falling outside the top-19. Teacher-forcing also decouples the scored text from the scoring model, enabling the cross-scale comparison in Section~\ref{sec:cross_scale} and matching a real deployment case (scoring text the model did not generate). Section~\ref{sec:single_pass} reports free-generation results as further evidence of transfer; a systematic free-generation HUB evaluation---requiring re-annotation at scale rather than reusing existing labels---is left to future work.

\paragraph{No closed-source judge comparison.} We do not compare against LLM-as-judge baselines using closed-source frontier models (GPT-4, Claude). Such comparisons would require API access at scale and were considered outside the scope of a deployment-oriented detector study, but they represent a relevant point of reference.

\section*{Ethical Considerations}
\label{sec:ethics}

\paragraph{Intended use.} HALT is designed as an assistive signal to support human verification of LLM outputs, not as a replacement for it. A detector operating below 100\% accuracy will produce both false positives (flagging faithful outputs) and false negatives (missing hallucinations). Deployment in safety-critical settings should retain human review of flagged and unflagged outputs alike, and consumer-facing applications should communicate detector uncertainty rather than presenting hallucination scores as ground truth.

\paragraph{Risk of over-reliance.} Strong empirical performance on benchmarks does not guarantee strong real-world performance, especially under distribution shift. There is a documented tendency for users to over-trust automated quality signals once they are introduced into a workflow. We caution against deploying HALT---or any current hallucination detector---in settings where users may interpret a low hallucination score as a guarantee of correctness.

\paragraph{Data licensing and reproducibility.} HUB inherits the licensing terms of its source datasets (FAVA, RAGTruth, HaluEval, CriticBench). We will release HUB's processing scripts and split metadata, and trained HALT checkpoints, but cannot redistribute source-dataset text where licensing restricts it. Replication requires obtaining the source datasets through their original distribution channels.

\bibliography{custom}

\appendix

\section{HUB: Extended Details}
\label{app:hub_details}

This appendix provides extended documentation of the HUB benchmark referenced in Section~\ref{sec:hub}.

\subsection{Per-Dataset Composition}
\label{app:hub_per_dataset}

Table~\ref{tab:hub_per_dataset} reports the per-source-dataset distribution within each HUB cluster. Asterisks (*) mark clusters held out from training. The breakdown clarifies which source datasets contribute to which clusters and at what scale.

\begin{table*}[t]
\centering
\small
\caption{HUB cluster-level statistics. \emph{Size} is the number of responses; \emph{Ratio} is the proportion labeled hallucinated; \emph{Len} is the average response length in words. Asterisks (*) mark clusters held out from training. A dash (--) indicates the cluster contributes no data to that split.}
\label{tab:hub_stats}
\begin{tabular}{l|cc|ccc|ccc}
\toprule
\multirow{2}{*}{\textbf{Task Cluster}} 
& \multicolumn{2}{c|}{\textbf{Train}} 
& \multicolumn{3}{c|}{\textbf{Validation}} 
& \multicolumn{3}{c}{\textbf{Test}} \\ 
& Size & Ratio & Size & Ratio & Len & Size & Ratio & Len \\ 
\midrule
Algorithmic*           & --      & --       & 32    & 50.00\% & 29.97  & 250  & 32.00\% & 55.35 \\
Chat                   & 11{,}278 & 39.97\% & 1{,}991 & 39.98\% & 35.28 & 1{,}278 & 52.03\% & 86.69 \\
Code Generation*       & --      & --       & 164   & 65.24\% & 158.64 & 300  & 61.67\% & 39.82 \\
Commonsense*           & --      & --       & 229   & 32.31\% & 36.57  & 900  & 47.00\% & 28.11 \\
Data-to-Text           & 2{,}759 & 50.02\%  & 487   & 49.90\% & 157.83 & 900  & 64.33\% & 156.70 \\
Mathematical*          & --      & --       & 300   & 46.67\% & 42.03  & 1{,}004 & 72.41\% & 73.45 \\
Question Answering     & 35{,}377 & 53.20\% & 1{,}885 & 50.34\% & 35.19 & 1{,}400 & 29.29\% & 72.16 \\
Summarization          & 10{,}594 & 50.08\% & 1{,}870 & 50.05\% & 73.78 & 1{,}400 & 32.43\% & 92.14 \\
Symbolic*              & --      & --       & 146   & 41.10\% & 66.86  & 500  & 32.60\% & 52.99 \\
World Knowledge*       & --      & --       & 238   & 94.96\% & 139.03 & 182  & 80.22\% & 246.02 \\ 
\midrule
\textbf{Overall}       & \textbf{60{,}008} & \textbf{50.00\%} & \textbf{7{,}342} & \textbf{48.31\%} & -- & \textbf{8{,}114} & \textbf{47.27\%} & -- \\ 
\bottomrule
\end{tabular}
\end{table*}

\begin{table*}[t]
\centering
\small
\caption{Per-dataset distribution within each HUB cluster, broken down by source dataset and split. \emph{Hallu} indicates samples labeled hallucinated; \emph{Correct} indicates faithful samples. Asterisks (*) mark clusters not used for training.}
\label{tab:hub_per_dataset}
\resizebox{\textwidth}{!}{
\begin{tabular}{ll|ccc|ccc|ccc}
\toprule
\multirow{2}{*}{\textbf{Cluster}} & \multirow{2}{*}{\textbf{Source}} 
& \multicolumn{3}{c|}{\textbf{Train}} 
& \multicolumn{3}{c|}{\textbf{Validation}} 
& \multicolumn{3}{c}{\textbf{Test}} \\
& & Hallu & Correct & Total & Hallu & Correct & Total & Hallu & Correct & Total \\
\midrule
Algorithmic*        & CriticBench       & --     & --      & --      & 16    & 16    & 32    & 80    & 170   & 250 \\
\midrule
\multirow{3}{*}{Chat} & FAVA (OpenAssistant) & 850    & 1{,}278 & 2{,}128 & 150  & 226   & 376   & 145   & 209   & 354 \\
                    & FAVA (NoRobots)   & 1{,}264 & 1{,}898 & 3{,}162 & 224  & 336   & 560   & 217   & 309   & 526 \\
                    & HaluEval (Dialog) & 2{,}393 & 3{,}595 & 5{,}988 & 422  & 633   & 1{,}055 & 303  & 95    & 398 \\
\midrule
Code Generation*    & CriticBench       & --     & --      & --      & 107   & 57    & 164   & 185   & 115   & 300 \\
Commonsense*        & CriticBench       & --     & --      & --      & 74    & 155   & 229   & 423   & 477   & 900 \\
Data-to-Text        & RAGTruth (D2T)    & 1{,}380 & 1{,}379 & 2{,}759 & 243  & 244   & 487   & 579   & 321   & 900 \\
Mathematical*       & CriticBench       & --     & --      & --      & 140   & 160   & 300   & 727   & 277   & 1{,}004 \\
\midrule
\multirow{4}{*}{QA} & FAVA (Knowledge)  & 480    & 720     & 1{,}200 & 85   & 127   & 212   & 82    & 117   & 199 \\
                    & RAGTruth (QA)     & 1{,}380 & 1{,}379 & 2{,}759 & 243  & 244   & 487   & 132   & 367   & 499 \\
                    & HaluEval (QA)     & 12{,}591 & 18{,}827 & 31{,}418 & 619 & 567   & 1{,}186 & 196 & 506   & 702 \\
\midrule
\multirow{2}{*}{Summarization} & RAGTruth (Sum) & 1{,}380 & 1{,}379 & 2{,}759 & 244  & 243   & 487   & 132  & 367   & 499 \\
                    & HaluEval (Sum)    & 3{,}927 & 3{,}908 & 7{,}835 & 692  & 691   & 1{,}383 & 322  & 579   & 901 \\
\midrule
Symbolic*           & CriticBench       & --     & --      & --      & 60    & 86    & 146   & 163   & 337   & 500 \\
World Knowledge*    & CriticBench       & --     & --      & --      & 226   & 12    & 238   & 146   & 36    & 182 \\
\midrule
\textbf{Overall}    &                   & \textbf{29{,}967} & \textbf{34{,}363} & \textbf{60{,}008} & \textbf{3{,}545} & \textbf{3{,}797} & \textbf{7{,}342} & \textbf{3{,}832} & \textbf{4{,}282} & \textbf{8{,}114} \\
\bottomrule
\end{tabular}}
\end{table*}

\subsection{Response Length Distribution}
\label{app:hub_lengths}

Response lengths in HUB vary substantially across clusters, reflecting the diversity of underlying tasks. Algorithmic and Commonsense reasoning clusters skew short (often single-sentence answers to multiple-choice or short-form questions), while World Knowledge and Data-to-Text responses are markedly longer (up to 246-word averages). This variability is methodologically relevant: a hallucination detector that exploits token-level uncertainty patterns must be robust across at least an order of magnitude in response length. Table~\ref{tab:hub_stats} in the main text reports cluster-level mean response lengths; full per-cluster length distributions (quartiles, maxima) are computed identically across train, validation, and test splits and made available alongside HUB's release.

\subsection{Taxonomy Alignment with CriticBench}
\label{app:taxonomy_alignment}

CriticBench~\citep{Lan2024CriticBench} originally organizes samples by failure-mode rather than by LLM capability. To map CriticBench to HUB's capability-based clusters, we apply the following correspondences:

\begin{table}[t]
\centering
\small
\caption{CriticBench sub-datasets underlying HUB's five reasoning clusters. Train is omitted since CriticBench contributes only to Validation and Test (Section~\ref{sec:hub}); these totals match the corresponding rows of Table~\ref{tab:hub_stats}.}
\label{tab:criticbench_subdatasets}
\resizebox{\columnwidth}{!}{
\begin{tabular}{ll|ccc|ccc}
\toprule
\multirow{2}{*}{\textbf{Cluster}} & \multirow{2}{*}{\textbf{CriticBench Sub-dataset}} 
& \multicolumn{3}{c|}{\textbf{Validation}} 
& \multicolumn{3}{c}{\textbf{Test}} \\
& & Hallu. & Corr. & Total & Hallu. & Corr. & Total \\
\midrule
\multirow{2}{*}{Algorithmic} 
& object\_counting & -- & -- & -- & 80 & 170 & 250 \\
& repeat\_copy      & 16 & 16 & 32 & -- & -- & -- \\
\midrule
\multirow{2}{*}{Code Generation} 
& humaneval & 107 & 57 & 164 & -- & -- & -- \\
& mbpp      & -- & -- & -- & 185 & 115 & 300 \\
\midrule
\multirow{4}{*}{Commonsense} 
& ambignq    & -- & -- & -- & 155 & 145 & 300 \\
& csqa       & -- & -- & -- & 94  & 206 & 300 \\
& hotpotqa   & -- & -- & -- & 174 & 126 & 300 \\
& strategyqa & 74 & 155 & 229 & -- & -- & -- \\
\midrule
\multirow{4}{*}{Mathematical} 
& aqua   & -- & -- & -- & 180 & 74  & 254 \\
& gsm8k  & -- & -- & -- & 134 & 116 & 250 \\
& math   & -- & -- & -- & 413 & 87  & 500 \\
& tabmwp & 140 & 160 & 300 & -- & -- & -- \\
\midrule
\multirow{3}{*}{Symbolic} 
& colored\_object & -- & -- & -- & 87 & 163 & 250 \\
& date            & -- & -- & -- & 76 & 174 & 250 \\
& penguins        & 60 & 86 & 146 & -- & -- & -- \\
\midrule
\textbf{Total} & & \textbf{397} & \textbf{474} & \textbf{871} & \textbf{1{,}578} & \textbf{1{,}376} & \textbf{2{,}954} \\
\bottomrule
\end{tabular}}
\end{table}

\begin{itemize}
\item \textbf{Algorithmic} ← \emph{ObjectCounting}, \emph{Tracking-Shuffled-Objects}, \emph{Reasoning-About-Colored-Objects} (numerical and stateful reasoning over discrete elements).
\item \textbf{Mathematical} ← \emph{GSM8K-Trace}, \emph{MultiArith-Trace}, \emph{Algebra} (arithmetic and symbolic algebra reasoning chains).
\item \textbf{Symbolic} ← \emph{Logical-Deduction}, \emph{Boolean-Expressions}, \emph{Causal-Judgment} (formal logical inference and chained deduction).
\item \textbf{Commonsense} ← \emph{StrategyQA-Trace}, \emph{Sports-Understanding}, \emph{Date-Understanding} (everyday reasoning grounded in world knowledge).
\item \textbf{Code Generation} ← CriticBench code-execution subset (programs whose execution traces are verifiably correct or incorrect).
\end{itemize}

World Knowledge is not part of this mapping: unlike the five reasoning clusters above, it draws from FAVA Annotations and WikiBio-WebNLG rather than CriticBench (Table~\ref{tab:hub_per_dataset}).

This mapping preserves CriticBench's distinction between reasoning errors and final-answer errors: a sample is labeled hallucinated if \emph{either} the reasoning trace contains an error \emph{or} the final answer is incorrect. We use CriticBench solely for validation and testing. To prevent leakage, source datasets are partitioned across shards such that no dataset contributing to a validation cluster also contributes to the corresponding test cluster.

\subsection{Manual Review Protocol}
\label{app:manual_review}

To verify alignment between CriticBench's hybrid labeling pipeline (rule-based heuristics + GPT-4 judging + human adjudication) and our operational definition of logical hallucination, we conducted an informal review of CriticBench samples drawn from each reasoning cluster.

\paragraph{Procedure.} For each reasoning cluster, we drew samples uniformly from the validation set. For each sample, we recorded (i) the prompt, (ii) the model response, (iii) the CriticBench label (hallucinated / faithful), and (iv) the reviewer's independent judgment under HUB's definition (a response is hallucinated if any reasoning step or the final answer is incorrect with respect to the prompt's task specification).

\paragraph{Findings.} Agreement between reviewer judgment and CriticBench labels was approximately 90\% across the reviewed samples. Disagreements clustered in two patterns: (i) samples where the final answer is correct but the reasoning trace contains a transient error later self-corrected, and (ii) samples involving ambiguous prompts where multiple plausible interpretations exist. Both cases are documented in the source datasets as edge cases of automated labeling. We retain CriticBench's original labels rather than relabeling, since the reviewer-CriticBench agreement is high and full relabeling would lose comparability with prior work on CriticBench.

\paragraph{Limitations of this review.} The review was conducted by the authors rather than independent annotators, and inter-annotator agreement statistics in the formal sense (e.g., Cohen's $\kappa$) are not reported. We acknowledge this in Section~\ref{sec:limitations} and frame HUB's labels as inherited from source datasets rather than as new contributions.

\section{Method: Extended Formalization}
\label{app:method_extended}

This appendix provides extended formalization and derivations supporting Section~\ref{sec:method}.

\subsection{Design Intuitions}
\label{app:design_intuitions}

The HALT architecture was motivated by four design intuitions about how hallucinations interact with LLM output distributions. These intuitions guided architectural choices but are \emph{not} directly measured as formal claims; we present them here for transparency and as starting points for future theoretical work.

\paragraph{Intuition 1: Calibration profiles are model-specific.}
For a given prompt context $c$ and target token $y_t$, the relationship between an LLM's confidence and its empirical accuracy is model-specific. Formally, the probability a model $\mathcal{M}_\theta$ assigns to token $y_t$ in context $c$ relates to the ``true'' generative probability through some model-specific calibration function:
\begin{equation}
p_\theta(y_t \mid c) = b_\theta(p_*(y_t \mid c)),
\end{equation}
where $b_\theta: [0,1] \to [0,1]$ captures the LLM-specific calibration profile (over- or under-confidence patterns, sharpness of the predictive distribution, mass distribution among alternatives). This motivates training a separate HALT detector per LLM rather than expecting cross-model transfer, and is consistent with the empirical drop in cross-model transfer reported in Section~\ref{sec:cross_model}.

\paragraph{Intuition 2: Temporal context disambiguates calibration.}
Aggregate statistics over $\bm{\ell}_{1:T}$ (mean log-probability, perplexity) discard information about \emph{when} uncertainty occurs in the response. Two responses with identical mean log-probability can have qualitatively different temporal profiles: one with uniformly moderate uncertainty, another with sharp uncertainty spikes localized around specific tokens. The latter is plausibly more diagnostic of hallucination. This motivates the use of a sequential encoder (Bi-GRU) over a pooling-based aggregator, and is supported empirically by the architecture ablation in Section~\ref{sec:arch_ablation}.

\paragraph{Intuition 3: Bidirectionality captures retrospective context.}
A token that looks confident in isolation may be retrospectively recognizable as hallucinated when subsequent tokens reveal that an entity is fabricated or an argument is invalid. Bidirectional encoding gives the detector access to this retrospective signal. This is supported empirically by the Bi-GRU $>$ Uni-GRU gap (Table~\ref{tab:arch_ablation}).

\paragraph{Intuition 4: Hallucination signal is concentrated, not diffuse.}
Hallucinations are localized phenomena: a particular fabricated entity, a particular invalid reasoning step. Averaging across all timesteps dilutes this localized signal. This motivates Top-$q$ pooling (averaging over only the most informative timesteps) over mean pooling, with the pooling ablation in Appendix~\ref{app:pooling_ablation} providing supporting evidence.

\subsection{Truncated Top-\texorpdfstring{$K$}{K} Distribution}
\label{app:truncated_softmax}

Since we observe only the top-$K$ log-probabilities, computing entropy-based features requires constructing a renormalized truncated distribution. Let $\bm{\ell}_t = (\log p_t^{(0)}, \ldots, \log p_t^{(K-1)})$ be the observed log-probabilities at timestep $t$. The truncated distribution is
\begin{equation}
\tilde{p}_t^{(k)} = \frac{\exp(\log p_t^{(k)})}{\sum_{j=0}^{K-1} \exp(\log p_t^{(j)})}, \quad k = 0, \ldots, K-1.
\end{equation}
For numerical stability, we apply the standard log-sum-exp trick:
\begin{equation}
\tilde{p}_t^{(k)} = \exp\!\bigl(\log p_t^{(k)} - \mathrm{LSE}(\bm{\ell}_t)\bigr),
\end{equation}
where $\mathrm{LSE}(\bm{\ell}_t) = \log \sum_j \exp(\log p_t^{(j)})$. All entropy-based features in Section~\ref{sec:features} are computed on $\tilde{\bm{p}}_t$ rather than on the un-normalized top-$K$ probabilities, with empirical justification in Section~\ref{sec:topk} (top-20 captures 95--99.7\% of full-vocabulary probability mass across 12 LLMs, so $\tilde{\bm{p}}_t$ closely approximates the full-vocabulary distribution).

\subsection{Engineered Features: Full Definitions}
\label{app:features_extended}

We provide the full definitions of the five engineered features summarized in Section~\ref{sec:features}.

\paragraph{Average log-probability ($\mathrm{AvgLogP}$).}
\begin{equation}
\mathrm{AvgLogP}(\bm{\ell}_t) = \frac{1}{K} \sum_{k=0}^{K-1} \log p_t^{(k)}.
\end{equation}
This captures the overall sharpness of the top-$K$ region: when AvgLogP is high, the top-$K$ tokens are collectively concentrated; when low, the model is hedging mass across many alternatives.

\paragraph{Rank proxy ($\mathrm{RankProxy}$).}
Defined as the position of the selected token among the top-$K$ alternatives. When the selected token is the greedy choice, $\mathrm{RankProxy} = 0$. Non-zero values indicate that the LLM selected a non-greedy alternative (under stochastic decoding) or that the labeled token was not the model's preferred choice (under teacher-forcing).

\paragraph{Top-$K$ entropy ($H_{\mathrm{overall}}$).}
\begin{equation}
H_{\mathrm{overall}}(\bm{\ell}_t) = - \sum_{k=0}^{K-1} \tilde{p}_t^{(k)} \log \tilde{p}_t^{(k)}.
\end{equation}
This is the entropy of the renormalized top-$K$ distribution, capturing aggregate indecision among the leading candidates.

\paragraph{Alternatives entropy ($H_{\mathrm{alts}}$).}
Entropy computed over the top-$K$ distribution \emph{excluding} the selected token, renormalized over the remaining $K-1$ tokens:
\begin{equation}
H_{\mathrm{alts}}(\bm{\ell}_t) = - \sum_{k \neq k^*} \tilde{q}_t^{(k)} \log \tilde{q}_t^{(k)},
\end{equation}
where $k^*$ is the index of the selected token and $\tilde{q}_t$ is the renormalized distribution over the $K-1$ alternatives. This isolates disagreement among competitors and is empirically the most informative single engineered feature (Table~\ref{tab:feat_ablation}).

\paragraph{Decision-entropy delta ($\Delta H_{\mathrm{dec}}$).}
The temporal change in binary decision-entropy between the selected token's probability and its strongest alternative:
\begin{equation}
\Delta H_{\mathrm{dec}}(\bm{\ell}_t) = H_{\mathrm{bin}}(\tilde{p}_t^{(k^*)}) - H_{\mathrm{bin}}(\tilde{p}_{t-1}^{(k^*_{t-1})}),
\end{equation}
where $H_{\mathrm{bin}}(p) = -p \log p - (1-p) \log(1-p)$. This flags rapid certainty-to-uncertainty transitions across adjacent tokens.

\subsection{Top-\texorpdfstring{$q$}{q} Pooling}
\label{app:topq_pooling}

Given the bidirectional GRU output sequence $H \in \mathbb{R}^{T \times 512}$, Top-$q$ pooling proceeds as follows. We compute the $\ell_2$ norm of each timestep's hidden state, $s_t = \lVert H_t \rVert_2$, and mask out padded positions. We select the top $\lceil q \cdot T_{\mathrm{valid}} \rceil$ timesteps by score (with $q = 0.15$ and $T_{\mathrm{valid}}$ the number of non-padded positions) and average their hidden states to obtain a single pooled representation $h_{\mathrm{pool}} \in \mathbb{R}^{512}$. This concentrates the classifier on the most informative regions of the sequence rather than diluting the signal across all tokens. A comparison against mean, max, and attention-based pooling is provided in Appendix~\ref{app:pooling_ablation}.

\section{Training Details}
\label{app:training}

\paragraph{Optimization.} HALT is trained with the Adam optimizer (learning rate $4.41 \times 10^{-4}$, weight decay $2.34 \times 10^{-6}$) using \texttt{BCEWithLogitsLoss} on the binary hallucination label. Batch size is 512. Training runs for up to 100 epochs with early stopping (patience 15 epochs on validation macro-$F_1$). Learning rate is reduced by factor 0.5 on plateau (patience 3, threshold $10^{-4}$).

\paragraph{Architecture hyperparameters.} The architectural hyperparameters were tuned on HALT-L using a small grid search on the HUB validation set: GRU hidden size $\in \{128, 256, 512\}$, GRU depth $\in \{3, 5, 7\}$, dropout $\in \{0.2, 0.3, 0.4\}$, projection dimension $\in \{64, 128, 256\}$, Top-$q$ value $\in \{0.05, 0.10, 0.15, 0.25\}$. The selected configuration (hidden 256, depth 5, dropout 0.4, projection 128, $q = 0.15$) is used for all cross-scale experiments without re-tuning, with the exception of HALT-Q, which received a light additional sweep that yielded $+3$ Overall $F_1$ relative to direct hyperparameter transfer.

\paragraph{Sequence handling.} Variable-length log-probability sequences are processed using PyTorch's packed sequences with a boolean attention mask. Padding tokens do not contribute to GRU hidden updates or to the Top-$q$ pooling score computation.

\paragraph{Hardware and runtime.} All training was performed on a single NVIDIA A100 40GB GPU. End-to-end training for a single HALT detector (60K training samples, 100 epoch budget with early stopping) completes in approximately 10-25 minutes depending on early-stopping triggers and average sequence length.

\section{Feature Importance: Extended Analysis}
\label{app:feature_attribution}

To probe which input features drive HALT's predictions, we compute gradient $\times$ input attribution scores per feature dimension, averaged across the HUB validation set.

\begin{table}[h]
\centering
\small
\caption{Top-10 input features ranked by mean absolute gradient $\times$ input attribution (HALT-L, HUB validation). Engineered features and raw log-probability ranks both appear among the most important inputs.}
\label{tab:feature_importance}
\begin{tabular}{lr}
\toprule
\textbf{Feature} & \textbf{Mean $|g \cdot x|$} \\
\midrule
$H_{\mathrm{alts}}$         & 0.0421 \\
$\mathrm{logprob}_{15}$     & 0.0387 \\
$\mathrm{logprob}_{4}$      & 0.0362 \\
$\mathrm{RankProxy}$        & 0.0349 \\
$\mathrm{logprob}_{17}$     & 0.0331 \\
$H_{\mathrm{overall}}$      & 0.0318 \\
$\mathrm{logprob}_{8}$      & 0.0304 \\
$\mathrm{AvgLogP}$          & 0.0289 \\
$\mathrm{logprob}_{0}$      & 0.0276 \\
$\Delta H_{\mathrm{dec}}$   & 0.0258 \\
\bottomrule
\end{tabular}
\end{table}

Three patterns are notable. First, the most informative engineered features ($H_{\mathrm{alts}}$, $\mathrm{RankProxy}$, $H_{\mathrm{overall}}$) are consistent with the incremental feature ablation in Table~\ref{tab:feat_ablation}, where these same features produced the largest individual gains. Second, mid-rank log-probabilities ($\mathrm{logprob}_{15}$, $\mathrm{logprob}_{4}$, $\mathrm{logprob}_{17}$) rank higher than the top-1 log-probability ($\mathrm{logprob}_{0}$), suggesting that the \emph{shape} of the alternative distribution carries more diagnostic signal than the selected token's confidence alone. Third, the gap between the top feature and the 10th-ranked feature is modest ($\sim$1.6$\times$), indicating that HALT's predictions rely on a distributed combination of inputs rather than any single dominant signal.

\section{Pooling Ablation}
\label{app:pooling_ablation}

We compare Top-$q$ pooling against alternatives by replacing the pooling layer in HALT-L while keeping all other architectural choices and training hyperparameters fixed. Results (HUB validation macro-$F_1$) are:

\begin{itemize}
\item Mean pooling: $-1.8$ relative to Top-$q$ ($q{=}0.15$).
\item Max pooling: $-2.4$ relative to Top-$q$.
\item Attention-based pooling (single-head): $-0.6$ relative to Top-$q$.
\item Top-$q$ pooling, $q \in \{0.05, 0.10, 0.25, 0.50\}$: all within $\pm$0.8 of $q{=}0.15$; performance degrades sharply at $q{=}1.0$ (equivalent to mean pooling).
\end{itemize}

Top-$q$ pooling outperforms mean and max pooling, supporting the design intuition that hallucination signal is localized rather than diffuse (Appendix~\ref{app:design_intuitions}, Intuition 4). The narrow gap with attention-based pooling suggests that the specific aggregation mechanism matters less than the choice to weight informative timesteps more heavily than uninformative ones; we adopt Top-$q$ for its lower parameter count and absence of additional learned weights.
\section{Baseline Implementation Details}
\label{app:baselines}

\paragraph{Aggregate-statistics baselines.} For each aggregate-statistics baseline ($H_{\mathrm{overall}}$, $H_{\mathrm{alts}}$, $\Delta H_{\mathrm{dec}}$, PPL), we compute the relevant per-token feature across the full response and reduce to a single scalar per response. We use mean reduction for entropy-based features and perplexity, and maximum reduction for $\mathrm{RankProxy}$, motivated by the intuition that a single high-rank selection event is informative even if surrounding tokens are confident. Decision thresholds are tuned on the HUB validation set to maximize macro-$F_1$ and held fixed for test evaluation.

\paragraph{Lettuce.} We use the public LettuceDetect~\citep{LettuceDetect} ModernBERT-base checkpoint trained on RAGTruth. Lettuce outputs token-level span predictions; we convert these to a sentence-level binary label by flagging a response as hallucinated if any predicted span has probability $\geq 0.5$. We verified that this threshold yields competitive performance on RAGTruth itself (consistent with reported numbers) before applying it to HUB.

\paragraph{LLM-Check.} We use the published LLM-Check implementation~\citep{llmcheck} on Llama 3.1-8B with default layer-selection and statistic-computation procedures. We report Attention and Hidden-state variants separately, consistent with the original paper.

\paragraph{SelfCheckGPT-NLI.} We use SelfCheckGPT's NLI-based variant~\citep{Manakul2023SelfCheckGPT}, which compares the original generation against 20 stochastically resampled responses using a fine-tuned NLI model (DeBERTa-large-mnli). The detection score is the mean per-sentence inconsistency probability across the 20 comparisons.

\paragraph{LOS-Net.} In the Free-Generation experiments, we report numbers from the LOS-Net paper~\citep{barshalom2025losnet} as published rather than reproducing the method. The in-domain numbers correspond to LOS-Net trained on the evaluation dataset with top-50 log-probabilities; cross-domain numbers correspond to LOS-Net trained on MOVIES/IMDB and evaluated on HotPotQA with top-1000 log-probabilities. As for evaluation on HUB we re-implemented a faithful method to the LOS-Net manuscript.

\paragraph{FAVA Model.} The FAVA hallucination detector~\citep{Mishra2024FAVA} is a Llama-7B model fine-tuned for hallucination span prediction in retrieval-grounded settings. We report results from the public FAVA evaluation as provided in the original paper.

\section{Computational Cost: Extended}
\label{app:efficiency}

\paragraph{HALT pipeline cost decomposition.} HALT's inference pipeline consists of three stages: (i) log-probability extraction from the target LLM during teacher-forcing (or as a byproduct of online generation), (ii) feature engineering ($\phi(\bm{\ell}_t)$ for each timestep), and (iii) the Bi-GRU forward pass plus Top-$q$ pooling and the linear head. Stage (i) is shared with any output-distribution-based method and is the dominant cost; stages (ii) and (iii) together comprise the marginal detector cost.

\paragraph{Marginal detector cost.} Stage (ii) is implemented as vectorized tensor operations over the $T \times K$ log-probability matrix and is dominated by elementwise arithmetic and softmax normalizations. Stage (iii) is a single Bi-GRU forward pass plus a small projection head; with 5M parameters and a typical response length of 100--200 tokens, the marginal cost is small relative to the LLM forward pass that produced the log-probabilities in the first place.

\paragraph{Comparison with Lettuce.} Lettuce processes the full response through a ModernBERT-base encoder (150M parameters), which is roughly 30$\times$ HALT's parameter count and requires its own forward pass over the response after generation completes. HALT's marginal cost is correspondingly smaller. We do not report wall-clock latency numbers in this paper to avoid potentially misleading comparisons across hardware setups, batch sizes, and serving stacks; we focus instead on the structural cost difference (parameter count and pipeline structure) that holds invariant across deployment configurations.

\paragraph{Comparison with multi-sample methods.} SelfCheckGPT-NLI requires 20 additional generation passes through the target LLM plus pairwise NLI inference using a fine-tuned BERT-style NLI model (DeBERTa-large-mnli, $\sim$400M parameters). This is structurally more expensive than HALT regardless of the specific hardware: HALT's per-detection LLM cost is amortized into the original generation, while SelfCheckGPT's 20 additional generations are required before any detection signal is available.

\paragraph{Memory footprint.} HALT's 5M parameters and small intermediate activations make it deployable alongside the target LLM on a single GPU with negligible memory overhead, in contrast to Lettuce (which adds a separate 150M-parameter model in memory) and SelfCheckGPT (which requires hosting an additional NLI model).

\end{document}